\documentclass[sigconf, nonacm]{acmart}

\newcommand\vldbpagestyle{plain} 
\usepackage{subcaption}
\usepackage{multirow}
\usepackage{enumitem}
\usepackage{graphicx}
\usepackage[linesnumbered,vlined,ruled,noend]{algorithm2e}
\usepackage{xcolor}
\usepackage{color, soul}

\begin{document}

\title{Efficient Graph Neural Network Inference at Large Scale } 


\author{Xinyi Gao}
\affiliation{
\institution{The University of Queensland}
}
\email{xinyi.gao@uq.edu.au}

\author{Wentao Zhang}
\affiliation{
  \institution{Mila - Québec AI Institute \\HEC Montréal}  
}
\email{wentao.zhang@mila.quebec}

\author{Yingxia Shao}
\affiliation{
\institution{Beijing University of Posts and Telecommunications}
}
\email{shaoyx@bupt.edu.cn}

\author{Quoc Viet Hung Nguyen}
\affiliation{
  \institution{Griffith University}  
}
\email{henry.nguyen@griffith.edu.au}

\author{Bin Cui}
\affiliation{
  \institution{Peking University}
}
\email{bin.cui@pku.edu.cn}

\author{Hongzhi Yin}
\affiliation{
  \institution{The University of Queensland}
}
\email{h.yin1@uq.edu.au}







\begin{abstract}
Graph neural networks (GNNs) have demonstrated excellent performance in a wide range of applications.
However, the enormous size of large-scale graphs hinders their applications under real-time inference scenarios. 
Although existing scalable GNNs leverage linear propagation to preprocess the features and accelerate the training and inference procedure, these methods still suffer from scalability issues when making inferences on  unseen nodes, as the feature preprocessing requires the graph is known and fixed. To speed up the inference in the inductive setting, we propose a novel adaptive propagation order approach that generates the personalized propagation order for each node based on its topological information. This could successfully avoid the redundant computation of feature propagation. Moreover, the trade-off between accuracy and inference latency can be flexibly controlled by simple hyper-parameters to match different latency constraints of application scenarios. To compensate for the potential inference accuracy loss, we further propose Inception Distillation to exploit the multi-scale reception information and improve the inference performance.  Extensive experiments are conducted on four public datasets with different scales and characteristics, and the experimental results show that our proposed inference acceleration framework outperforms the SOTA graph inference acceleration baselines in terms of both accuracy and efficiency. In particular, the advantage of our 
proposed method is more significant on larger-scale datasets, and our framework achieves $75\times$ inference speedup on the largest Ogbn-products dataset.


\end{abstract}

\maketitle

\pagestyle{\vldbpagestyle}


\section{Introduction}
Developing a graph neural network (GNN) for very large graphs has drawn increasing attention due to the powerful expressiveness of GNNs and their enormous success in many industrial applications~\citep{wu2020graph,DBLP:conf/icde/HuG0J19,DBLP:conf/sigmod/Vretinaris0EQO21}. 
Although, GNNs provide a universal framework to tackle various down-streaming tasks, performing the model on large-scale industrial graphs suffers from heavy computation and high latency. 
This severely limits the application to latency-sensitive scenarios.
For example, recommender systems designed for streaming sessions must completely perform real-time inference on user-item interaction graphs \citep{chandramouli2011streamrec, wu2019session, qiu2020gag, wang2020next}. The fraud and spam detection tasks require millisecond-level inference on the million-scale graph to identify the malicious users and avoid the property loss of the victim users \citep{wang2019semi, 101145, liu2018heterogeneous}. In some computer vision applications, GNNs are designed for 3D point clouds data and deployed on edge devices such as self-driving cars to perform object detection or semantic segmentation tasks \citep{qi20173d, shi2020point, landrieu2018large}. In such scenarios, real-time inference response is essential.

The root cause for the heavy computation and high latency of GNNs is the neighbor explosion problem.
Generally, most GNNs adopt the message-passing pipeline and leverage the feature propagation and transformation procedures to construct the model. Through executing $k$ times of feature propagation, the $k$-th order propagated features can capture the node information from $k$-hop neighborhoods. Especially in large-scale and sparsely labeled graphs, multiple layers of propagation are needed to aggregate enough label information from distant neighbors according to the message-passing pipeline \citep{hu2020ogb, zhang2022model, zhang2021rod, zeng2021decoupling, trung2020adaptive}. However, as the order of propagation layers increases, the number of supporting nodes grows exponentially. This directly incurs the high computation cost of feature propagation.

To mitigate the expensive computation resulted from feature propagation, several linear propagation-based GNNs \citep{wu2019simplifying,sign_icml_grl2020,zhang2022pasca,chen2020scalable,zhang2022nafs,zhu2020simple,zhang2022graph}, e.g., SGC, were proposed to remove the non-linearity among feature propagation and aggregate node features during the preprocessing procedure. 
Instead of performing feature propagation during each training epoch, this time-consuming process is only executed once in linear propagation-based GNNs. 
As a result, the time complexity of model training is significantly reduced, and the training of these models scales well with graph size. 
However, linear propagation-based GNNs still struggle with efficient inference at scale because the preprocessing of feature propagation is based on the premise that the graph is known and fixed. 
This strong premise severely limits real-world applications, and more practical scenarios require inference on unseen nodes, where feature propagation has to be executed online. In addition, these existing methods adopt a fixed propagation order for all nodes. Due to the complex topological structures, the fixed propagation order restricts the flexibility of exploiting the multi-scale reception fields and also tends to over-smooth the high-degree nodes, leading to wasted computation and performance degradation. 

To this end, we propose to reduce the redundant computation of feature propagation to further accelerate the inference of scalable GNNs. 
Specifically, we design a plug-and-play technique: Node-Adaptive Inference (NAI), which introduces node-wise adaptive propagation order (or propagation depth) to customize the propagation order for each node. 
By measuring the distance between the current feature and the stationary state, the smoothing status of the propagated feature is evaluated. Then we introduce simple global hyper-parameters to adaptively determine the propagation order for each node and efficiently trade off between inference latency and accuracy.
This provides a variety of inference options for users with different latency constraints. Moreover, we design a novel Inception Distillation module in NAI to exploit the multi-scale reception field information and mitigate performance degradation. With a more powerful supervision signal, NAI could accelerate the inference speed with a negligible performance drop.


The main contributions of this paper are summarized as follows:
\begin{itemize}[leftmargin=*]
  \item \textbf{New Scenario.} We focus on the inference speedup in a more real and challenging setting - graph-based inductive inference, where the ever-scalable GNNs also struggle with heavy online computation of feature propagation.
  \item \textbf{New Methodology.} Instead of using the fixed order of feature propagation as done in existing GNNs and other acceleration methods, we propose a novel adaptive propagation order approach that generates the personalized propagation order for each node based on its topological information. This could successfully avoid the redundant computation of feature propagation and mitigate the over-smoothing problem. Moreover, the trade-off between accuracy and inference latency can be flexibly controlled by simple hyper-parameters. To compensate for the potential inference accuracy loss, we further propose Inception Distillation to exploit the multi-scale reception information to improve the inference performance.
  \item \textbf{SOTA Performance.}   Extensive experiments are conducted on four public datasets with different scales and characteristics, and the experimental results show that our proposed efficient inference framework NAI  outperforms the SOTA graph inference acceleration baselines in terms of both accuracy and efficiency. In particular, the advantage of our NAI is more significant on larger-scale datasets, and NAI achieves $75\times$ inference speedup on the largest Ogbn-products dataset.
\end{itemize}


\section{Preliminary}
\subsection{Problem Formulation}
Given a graph $\mathcal{G}$ = ($\mathcal{V}$, $\mathcal{E}$) with $|\mathcal{V}| = n$ nodes and $|\mathcal{E}| = m$ edges, its node adjacency matrix and degree matrix are denoted as ${\mathbf{A}} \in \mathbb{R}^{n \times n}$ and $\mathbf{D} = \mathrm{diag}(d_1, d_2, ..., d_n$), where $d_i =  {\textstyle \sum_{v_j\in \mathcal{V}}\mathbf{A}_{i,j}} $ is the degree of node $v_i$. The adjacency matrix and degree matrix with self-loops are denoted as $\widetilde{\mathbf{A}} $ and $\widetilde{\mathbf{D}}$. The node feature matrix is $\mathbf{X} = \{\boldsymbol{x}_1, \boldsymbol{x}_2, ..., \boldsymbol{x}_n\}$ in which $\boldsymbol{x}_i\in\mathbb{R}^{f}$ represents the node attribute vector of $v_{i}$, and $\mathbf{Y} = \{\boldsymbol{y}_1, \boldsymbol{y}_2, ..., \boldsymbol{y}_l\}$ is the one-hot label matrix for classification task. 
In the inductive setting, the entire node set $\mathcal{V}$  is partitioned into training set $\mathcal{V}_{train}$ (including labeled set $\mathcal{V}_l$ and unlabeled set $\mathcal{V}_u$) and test set $\mathcal{V}_{test}$. GNNs are trained on $\mathcal{G}_{train}$ which only includes $\mathcal{V}_{train}$ and all edges connected to $v \in \mathcal{V}_{train}$. The evaluation is to test the performance  of trained GNNs on $\mathcal{V}_{test}$.

\subsection{Scalable Graph Neural Networks}
\textbf{GNNs} aim to learn node representation by using topological information and node attributes. The existing GNNs adopt the message-passing pipeline and construct models utilizing two processes: feature propagation and transformation. By stacking multiple layers, the $(k+1)$-th layer feature matrix $\mathbf{X}^{(k+1)}$ can be formulated as:
\begin{equation}
\begin{aligned}
    & \mathbf{X}^{(k+1)} =\delta\left(\hat{\mathbf{A}}\mathbf{X}^{(k)}\mathbf{W}^{(k)}\right),\\
    & \hat{\mathbf{A}}=\widetilde{\mathbf{D}}^{r-1}\widetilde{\mathbf{A}}\widetilde{\mathbf{D}}^{-r},
\end{aligned}    
\label{eq_GCN}
\end{equation}
where $\mathbf{W}^{(k)}$ is the model weights,  $\delta\left(\cdot\right)$ is the activation function and $\widetilde{\mathbf{D}}$ is the diagonal node degree matrix used to normalize $\widetilde{\mathbf{A}}$.
In each layer, $\hat{\mathbf{A}}$ propagates the information among neighbors, and $\mathbf{W}^{(k)}$ transforms the propagated features. Note that, $r \in [0, 1]$ is the convolution coefficient and could generalize Eq. (\ref{eq_GCN}) to various existing models. By setting $r=1$, 0.5 and 0, the convolution matrix $\hat{\mathbf{A}}$ represents the transition probability matrix $\widetilde{\mathbf{A}}\widetilde{\mathbf{D}}^{-1}$ \citep{chiang2019cluster, DBLP:conf/iclr/ZengZSKP20, hamilton2017inductive},
the symmetric normalization adjacency matrix $\widetilde{\mathbf{D}}^{-\frac{1}{2}}\widetilde{\mathbf{A}}\widetilde{\mathbf{D}}^{-\frac{1}{2}}$ \citep{gasteiger_predict_2019, DBLP:conf/iclr/KipfW17} and the reverse transition probability matrix $\widetilde{\mathbf{D}}^{-1}\widetilde{\mathbf{A}}$ \citep{DBLP:conf/icml/XuLTSKJ18}, respectively.

\textbf{Linear Propagation-based Scalable GNNs}. Although GNNs achieve excellent performance by executing multiple feature propagation and transformation processes, it was found that the aggregation of neighbor features (i.e., feature propagation) makes a major contribution to the performance of GNNs and plays a more important role \citep{wu2019simplifying}. Based on this finding, to improve the scalability of GNNs,  SGC \citep{wu2019simplifying} was proposed to decompose the two processes and remove feature transformations in the middle layers. It propagates the node features for $k$ times and then feeds $k$-th order propagated feature ${\mathbf{X}}^{(k)}=\hat{\mathbf{A}}^{k}\mathbf{X}$ to a linear model for classification. Benefiting from the linear propagation, SGC facilitates the precomputation of the feature matrix and successfully reduces the training time. 
 
 \begin{figure*}[t]
\setlength{\abovecaptionskip}{0.cm}
\includegraphics[width=\linewidth]{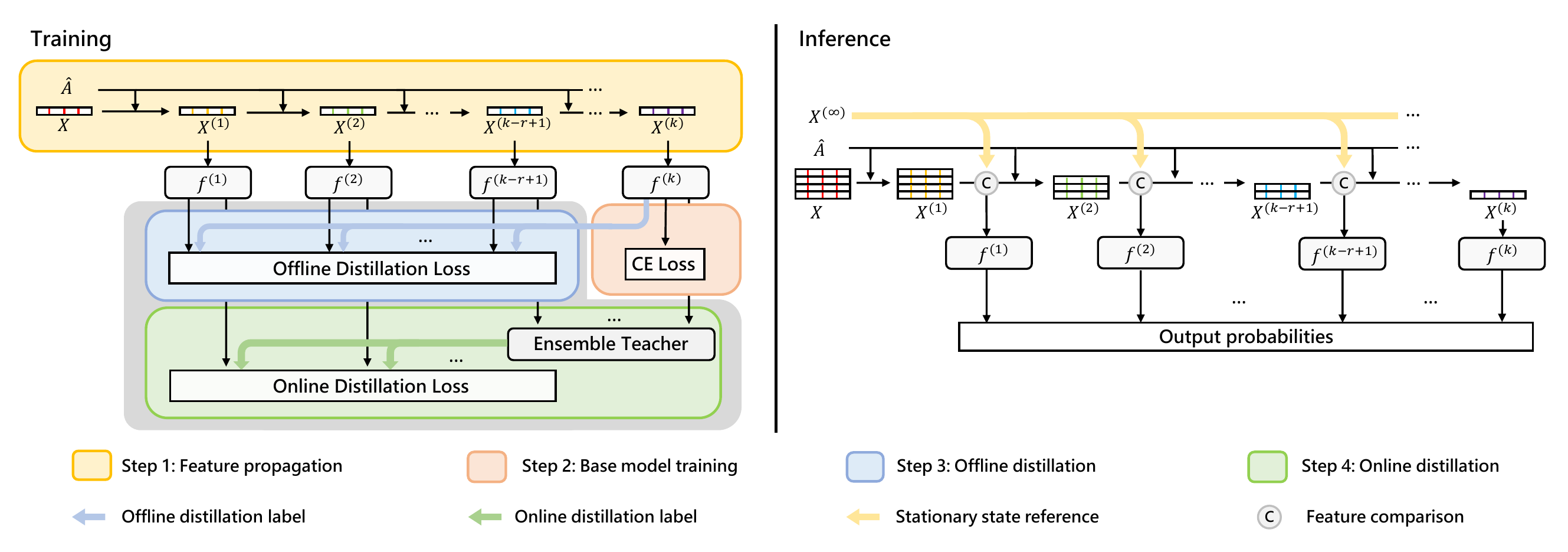}
\caption{The training and inference procedure for NAI. The training procedure (left) includes feature propagation, base model training and Inception Distillation. In inference procedure (right), the propagation order is adaptively controlled by comparing the propagated feature with the stationary state.}
\label{fig_main}
\end{figure*}

 Following SGC, more powerful scalable GNNs are designed by adopting linear propagation. For example, SIGN \citep{sign_icml_grl2020} proposes to transform propagated features in different orders by linear transformations, then concatenates them together to enhance the feature representation. The transformation objective can be represented as : ${\mathbf{X}}^{(0)}\mathbf{W}_0\left |  \right |  {\mathbf{X}}^{(1)}\mathbf{W}_1\left |  \right |  ...\left |  \right |  {\mathbf{X}}^{(k)}\mathbf{W}_k$, where $\left |  \right | \cdot \left |  \right |$ denotes concatenation operations and $\mathbf{W}_k$ are transformation matrixes. $\mathrm{S^2GC}$ \citep{zhu2020simple} averages propagated features in different orders to construct a simple spectral graph convolution: $\frac{1}{k} {\textstyle \sum_{l=0}^{k}}{ \mathbf{X}}^{(l)}$. 
 GAMLP\citep{zhang2022graph} combines propagated features in different orders by measuring the feature information gain and constructing the node-wise attention mechanism:${\textstyle \sum_{l=0}^{k}} T^{(l)}{\mathbf{X}}^{(l)}$, where $T^{(l)}$ are diagonal node-wise attention matrices. The non-parametric feature propagation used in these methods can successfully speed up the training procedure and transductive graph inference by preprocessing the propagated features. However, they cannot accelerate the graph inductive inference on unseen nodes, as the feature preprocessing requires the graph is known and fixed. 
\section{Method}
\subsection{Architecture Overview}
Figure \ref{fig_main} shows the overview of NAI for linear propagation-based GNNs. Without loss of generality, we deploy NAI on SGC as an example. For the training procedure, NAI employs Inception Distillation to compensate for the potential inference accuracy loss, which includes two steps: offline distillation and online distillation. Specifically, given the raw feature matrix $\mathbf{X}$, we first compute the propagated features in different orders ${\mathbf{X}}^{(l)}$, where $1 \le l \le k$. Then, with the largest reception field information and the best performance, the base model $f^{(k)}$ is trained with ${\mathbf{X}}^{(k)}$, and its knowledge is distilled into other $k-1$ classifiers by means of offline distillation. 
Besides the single-scale knowledge within $f^{(k)}$, we wish our model can capture multi-scale information of different-sized reception fields to help improve the inference performance. 
To this end, we introduce the self-attention mechanism to construct a more powerful teacher model for distillation. The predictions of $r$ enhanced classifiers are adaptively combined to provide the supervision signals, and both student and teacher are updated simultaneously according to the online distillation loss. 

As for the inference procedure, the distances between the propagated features and the stationary states are measured, and nodes with satisfied feature smoothness are inferred by well-trained classifiers. As a result, the personalized propagation order is adaptively generated for each node, avoiding the redundant computation of feature propagation and the over-smoothing risk.

\subsection{Inception Distillation}
For a scalable GNN including $k$ times propagation, the base model $f^{(k)}$ is trained with ${\mathbf{X}}^{\left(k\right)}$ by using Cross-Entropy (CE) loss between the predicted softmax outputs and the one-hot labels. 

\begin{equation}
\begin{aligned}
&\mathcal{L}^{(k)}=-\frac{1}{\left | \mathcal{V}_l \right | }  \sum_{v_i\in {\mathcal{V}_l}}\boldsymbol{y}_i\log \tilde{\boldsymbol{y}}^{(k)}_i,\\
&\tilde{\boldsymbol{y}}^{(k)}_i=\mathrm{softmax}({\boldsymbol{z}}^{\left(k\right)}_i),\\
&{\boldsymbol{z}}^{(k)}_i=f^{(k)}({\mathbf{X}}^{\left(k\right)}_i),
\end{aligned} 
\label{eq_initialloss}
\end{equation}
where $\boldsymbol{y}_i$ and $\tilde{\boldsymbol{y}}^{(k)}_i$ are one-hot label and classifier's softmax output of node $v_i$. Then, the knowledge of $f^{(k)}$ will be distilled in other student classifiers. We penalize the soft CE loss between the student’s softmax outputs against the teacher’s softmax outputs.
\begin{equation}
\begin{aligned}
&\mathcal{L}^{(l)}_{d}=-\frac{1}{\left | \mathcal{V}_{train} \right | } \sum_{v_i\in {\mathcal{V}_{train}}}\tilde{\boldsymbol{p}}^{(k)}_i\log \tilde{\boldsymbol{p}}^{(l)}_i,\\
&\tilde{\boldsymbol{p}}^{(l)}_i=\mathrm{softmax}(z^{(l)}_{i}/T),\\
&\tilde{\boldsymbol{p}}^{(k)}_i=\mathrm{softmax}(z^{(k)}_{i}/T),\\
&z^{(l)}_{i}=f^{(l)}({\mathbf{X}}^{\left(l\right)}_i),\\
\end{aligned} 
\label{eq_offlinekdloss}
\end{equation}where  $1\le l<k$. $T$ is the temperature, which controls how much to rely on the teacher’s soft predictions \citep{hinton2015distilling}. A higher temperature produces a more diverse probability distribution over classes. Besides $\mathcal{L}^{(l)}_{d}$, the node label provides another supervision signal for the student models, and the offline distillation loss $\mathcal{L}^{(l)}_{off}$ is constructed by jointly optimizing $\mathcal{L}^{(l)}_c$ and $\mathcal{L}^{(l)}_{d}$:
\begin{equation}
\begin{aligned}
&\mathcal{L}^{(l)}_{off}=(1-\lambda)\mathcal{L}^{(l)}_c+\lambda T^2\mathcal{L}^{(l)}_{d},\\
&\mathcal{L}^{(l)}_c=-\frac{1}{\left | \mathcal{V}_l \right | } \sum_{v_i\in {\mathcal{V}_l}}\boldsymbol{y}_i\log \tilde{\boldsymbol{y}}^{(l)}_i,\\
&\tilde{\boldsymbol{y}}^{(l)}_i=\mathrm{softmax}({\boldsymbol{z}}^{\left(l\right)}_i),\\
\end{aligned} 
\label{eq_offlineloss}
\end{equation} where $T^2$ is used to adjust the magnitudes of the gradients produced by knowledge distillation \citep{hinton2015distilling} and $\lambda \in [0, 1]$ is the  hyper-parameter that balances the importance of two losses.


With enhanced classifiers for propagated features in different orders, the ensemble teacher is built to preserve multi-scale reception signals. It is voted by $r$ classifiers and their predictions are combined as: 
\begin{equation}
\begin{aligned}
&\bar{\boldsymbol{z}}_{i}=\mathrm{softmax}(\sum_{l=k-r+1}^{k} w^{(l)}_{i}\tilde{\boldsymbol{y}}^{(l)}_{i}),\\
&w^{(l)}_{i}=\frac{exp(m^{(l)}_{i})}{\sum_{l=k-r+1}^{k} exp(m^{(l)}_{i})},\\
&m^{(l)}_{i}=\delta (\tilde{\boldsymbol{y}}^{(l)}_{i}s),
\end{aligned} 
\label{eq_onlineteacher}
\end{equation}where $\bar{\boldsymbol{z}}_{i}$ is the ensemble teacher prediction for node $v_i$ and $\delta\left(\cdot\right)$ is the activation function. $s\in \mathbb{R}^{f \times 1}$ is the weight vector which projects the logits into the subspace to measure self-attention scores. Scalars $m^{(l)}_{i}$ are normalized to weight the predictions $\tilde{\boldsymbol{y}}^{(l)}_{i}$. Then, the student model and the weight vector $s$ are optimized by minimizing the online distillation loss $\mathcal{L}^{(l)}_{on}$:

\begin{equation}
\begin{aligned}
&\mathcal{L}^{(l)}_{on}=(1-\lambda)\mathcal{L}^{(l)}_{c}+\lambda T^2\mathcal{L}^{(l)}_{e},\\
&\mathcal{L}^{(l)}_{e}=-\frac{1}{\left | \mathcal{V}_{train} \right | }\sum_{v_i\in {\mathcal{V}_{train}}}\bar{\boldsymbol{p}}_i\log \tilde{\boldsymbol{p}}^{(l)}_i,\\
&\bar{\boldsymbol{p}}_i=\mathrm{softmax}(\bar{\boldsymbol{z}}_{i}/T),\\
\end{aligned} 
\label{eq_onlineloss}
\end{equation}
where $1 \le l < k$. 
The ensemble teacher built from $r$ models will be updated simultaneously with students. By utilizing $\mathcal{L}^{(l)}_{off}$ and $\mathcal{L}^{(l)}_{on}$, Inception Distillation could capture comprehensive knowledge within multi-scale reception fields to improve the performance of each student classifier.

\subsection{Node-Adaptive Propagation}
In the inference procedure, we introduce the novel Node-Adaptive Propagation (NAP) module to generate personalized propagation order/depth for each node. Scalable GNNs propagate the information within $k$-hops neighbors by multiplying the $k$-th order normalized adjacency matrix by the feature matrix: $\hat{\mathbf{A}}^{k}\mathbf{X}$. This operation gradually smooths the node feature by neighbors, and with the growth of the order, the propagated node features within the same connected component will reach a stationary state \citep{li2018deeper}.  When $k\to \infty$, the stationary feature state $\mathbf{X^{\left(\infty\right)}}$ can be calculated as:
\begin{equation}
\begin{aligned}
&\mathbf{X}^{\left(\infty\right)}=\hat{\mathbf{A}}^{\left(\infty\right)}\mathbf{X},\\
&\mathbf{\hat{A}}^{\left(\infty\right)}_{i,j}=\frac{\left ( d_i+1 \right ) ^{r}\left (  d_j+1\right ) ^{1-r}}{2m+n},
\end{aligned}    
\label{eq_stationary}
\end{equation}where $\mathbf{\hat{A}}^{\left(\infty\right)}_{i,j}$ is the weight between nodes $v_i$ and $v_j$ , i.e., the element of $i$-th row and $j$-th column in $\mathbf{\hat{A}}^{\left(\infty\right)}$. $d_i$ and $d_j$ are node degrees for $v_i$ and $v_j$. $m$ and $n$ are the numbers of edges and nodes. $r$ is the convolution coefficient in Eq. (\ref{eq_GCN}).

With the definition of stationary feature state, the smoothness of node features can be well evaluated. Inspired by \cite{zhang2021node}, we use the distance between the propagated feature $\mathbf{X}^{\left(l\right)}_i$ and stationary feature $\mathbf{X}^{\left(\infty\right)}_i$ to measure the feature smoothness of node $v_i$, and the distance $\mathbf{d}^{\left(l\right)}_i$ is defined as Eq. (\ref{eq_distance}).
\begin{equation}
\begin{aligned}
\mathbf{d}^{\left(l\right)}_i = \; \left \| {\mathbf{X}}^{\left(l\right)}_i  - {\mathbf{X}}^{\left(\infty\right)}_i\right \|,
\end{aligned}    
\label{eq_distance}
\end{equation}
where $\left \| \cdot \right \|$ means $l_2$ norm. Then, different from existing GNNs that adopt $\mathbf{X}^{\left(k\right)}_i$ directly, the personalized propagation order for the node $v_i$ is generated according to the inference algorithm \ref{alg:Framwork}.




  \begin{algorithm}[htb]
  \caption{Working pipeline of NAP.}
  \LinesNumbered
  \label{alg:Framwork}
   \KwIn{$k$ classifiers, adjacent matrix $\mathbf{\widetilde{A}}$, degree matrix $\mathbf{\widetilde{D} }$, feature matrix $\mathbf{X}$, test set $\mathcal{V}_{test}$, propagation order $k$, threshold $T_s$, the minimum propagation order $T_{min}$ and the maximum propagation order $T_{max}$.}
    \KwOut{The prediction results of $\mathcal{V}_{test}$.}
    \For{\rm{batch} $\mathcal{V}_{b}$ \rm{in} $\mathcal{V}_{test}$}{
    Calculate the stationary feature state $\mathbf{X^{\left(\infty\right)}}$ for $\mathcal{V}_{b}$;\\
    Sample supporting nodes for $\mathcal{V}_{b}$;\\
    \For{$l=1$ \rm{to} $T_{max}$}{
    Calculate the propagated feature ${\mathbf{X}}^{(l)}$ for $\mathcal{V}_{b}$;\\
    \If{$l<T_{min}$}{
        Continue;\\}
    \ElseIf{$l<T_{max}$}{
    \For{$i=1$ \rm{to} $|\mathcal{V}_{b}|$}{
     Calculate the distance $\mathbf{d}^{\left(l\right)}_i$ between ${\mathbf{X}_i}^{(l)}$ and $\mathbf{X}_i^{\left(\infty\right)}$ for test node $v_i$;\\
        \If{$\mathbf{d}_i<T_{s}$}{
            Predict ${\mathbf{X}_i}^{(l)}$ by classifier $f^{(l)}$;\\
            Remove $v_i$ from $\mathcal{V}_{b}$;\\
            }
        \Else{
    Continue;\\}}}
    \Else{
    Predict $\mathcal{V}_{b}$ by classifier $f^{(l)}$;\\}
    }}
    \Return The prediction results for $\mathcal{V}_{test}$.
\end{algorithm}

To adapt NAP to different latency constraints and application scenarios, we introduce three simple hyper-parameters in the inference algorithm, i.e., $T_s$, $T_{min}$ and $T_{max}$. $T_s$ is used to control the smoothing effect. A larger $T_s$ indicates a weak smoothing effect and smaller propagation order is required. $T_{min}$ and $T_{max}$ are the minimum and the maximum propagation order, respectively. 
In line 2-3, $\mathbf{X^{\left(\infty\right)}}$ and supporting nodes are firstly derived according to $\mathcal{V}_{b}$ and $T_{max}$, where $1\le T_{max}\le k$. Then, the node features will be propagated $T_{min}$ times, where $1\le T_{min}\le T_{max}$ (line 5). After $T_{min}$ times propagation, features are compared with $\mathbf{X^{\left(\infty\right)}}$ and inferred by the classifier if the distances are smaller than $T_{s}$ (line 9-12). Until $l=T_{max}$, all left nodes will be classified by $f^{(T_{max})}$ and the prediction results for $\mathcal{V}_{b}$ are output (line 17-18). 
After deploying the model on the device, users can easily search the hyper-parameters that match the latency requirements and select the one that yields the highest validation accuracy for inference.

\subsection{Complexity Analysis}


\begin{table*}[pt]
\caption{{The computational complexities of scalable GNNs in the inductive setting. $n$, $m$ and $f$ are the number of nodes, edges, and feature dimensions, respectively. $k$ denotes the propagation order and $P$ is the number of layers in classifiers. $q$ is the averaged propagation order when adopting NAI.}}
\label{tab_complexity}
\resizebox{0.65\textwidth}{!}{
\begin{tabular}{l|llll}
\toprule
                   & SGC & ${\mathrm{S^2GC}}$ & SIGN & GAMLP \\ \hline
Vanilla & $\mathcal{O} (kmf+nf^2)$  &   $\mathcal{O} (kmf+knf+nf^2)$ &  $\mathcal{O} (kmf+kPnf^2)$      &   $\mathcal{O}(kmf+Pnf^2)$     \\
NAI &  $\mathcal{O}(qmf+nf^2)$  &   $\mathcal{O}(qmf+qnf+nf^2)$ &  $\mathcal{O}(qmf+qPnf^2)$       &   $\mathcal{O}(qmf+Pnf^2)$ \\ \bottomrule
\end{tabular}}
\end{table*}

Table \ref{tab_complexity} compares the computational complexity of four linear propagation-based GNNs and their complexity after deploying NAI in the inductive setting.
All computations include feature processing and classification, and we show the simplest version of GAMLP which utilizes the attention mechanism in the feature propagation. NAI could reduce the computation of feature processing by decreasing the propagation order $k$. Suppose $q$ is the average propagation order over all nodes when adopting NAI, the complexity for feature processing in SGC is decreased to $\mathcal{O} (qmf)$. This means that NAI can achieve stronger acceleration effects for graphs with large-scale edges and high feature dimensions under the same $q$. 
The classification complexity is $\mathcal{O} (nf^2)$, which is same as vanilla SGC. Similar results can be observed in ${\mathrm{S^2GC}}$ and GAMLP. For SIGN, it concatenates propagated features in different orders before the classification procedure, leading to the increase of feature dimension. 
As a result, the classification computation also decreases from $\mathcal{O}(kPnf^2)$ to $\mathcal{O}(qPnf^2)$ when applying NAI to SIGN. 
\section{Experiments}

\subsection{Experimental Settings}
\textbf{Datasets}. We evaluate our proposed method on four public datasets with different scales and characteristics, including: two citation networks (PubMed and Ogbn-arxiv) \citep{DBLP:conf/iclr/KipfW17, DBLP:conf/nips/HuFZDRLCL20}, a image network (Flickr) \citep{DBLP:conf/iclr/ZengZSKP20} and a product co-purchasing network (Ogbn-products) \citep{DBLP:conf/nips/HuFZDRLCL20}. In citation networks, papers from different topics are considered as nodes and the edges are citations among the papers. 
Flickr contains descriptions and properties of images and the node class is the image category. 
In Ogbn-products, the nodes representing products, and edges between two products indicate that the products are purchased together. 
The detailed descriptions of the datasets are provided in Table \ref{tab_data}.

\begin{table}[ht]
\caption{Datasets properties. $n$, $m$, $f$ and $c$ are the number of nodes, edges, feature dimensions and classes, respectively.}
\label{tab_data}
\resizebox{\linewidth}{!}
{\begin{tabular}{l|rrrrl}
\toprule
Dataset    & $n$ & $m$ & $f$ & $c$ & \#Train/Val/Test\\ \midrule
PubMed     & 19,717   & 44,338   & 500        & 3         & 60/500/1,000                  \\
Flickr     & 89,250   & 899,756   & 500        & 7        & 44k/22k/22k                  \\ 
Ogbn-arxiv    & 169,343   & 1,166,243   & 128        & 40        & 91k/30k/48k                  \\ 
Ogbn-products      & 2,449,029   & 123,718,280  & 100        & 47  &196k/39k/2,213k
\\ \bottomrule
\end{tabular}}
\end{table}


\noindent\textbf{Baselines}. We compare NAI with the vanilla base model and state-of-the-art methods designed for inference acceleration, which includes: (1) GLNN \citep{zhang2021graph}. Distill the knowledge from a deep GNN teacher to a simple MLP to eliminate the neighbor-fetching latency in GNN inference. Note that GLNN completely abandons the feature propagation to speed up the inference and can be seen as the extremely simplified case of NAI. (2) TinyGNN \citep{yan2020tinygnn}. Distill the knowledge from a deep GNN teacher to a single-layer GNN while exploiting the local structure information within peer nodes. (3) Quantization. Quantize model parameters from FP32 to INT8.

\noindent\textbf{Evaluation metrics}. The performance of each baseline is evaluated by five criteria, including the accuracy of the test set (ACC), averaged multiplication-and-accumulation operations per node (MACs), averaged feature processing MACs per node (FP MACs), averaged inference time per node (Time) and averaged feature processing time per node (FP time). Notice that MACs for NAI evaluates 4 procedures, including stationary state computation, feature propagation, distance computation and classification. Besides these procedures, the Time for NAI further contains the time of supporting node sampling. FP MACs and FP Time for NAI evaluate the feature propagation and distance computation procedure.

\noindent\textbf{Implementation and Settings}. Without loss of generality, we use the symmetric normalization adjacency matrix $\widetilde{\mathbf{D}}^{-\frac{1}{2}}\widetilde{\mathbf{A}}\widetilde{\mathbf{D}}^{-\frac{1}{2}}$ in all base models. For each method, the hyper-parameters used in experiments are searched by the grid search method or following the original papers, and we use the ADAM optimization algorithm to train all the models. The best propagation order $k$ for each dataset and base model is searched together with learning rate, weight decay, and dropout to get the highest performance. Specifically, the values for $k$, learning rate and weight decay are searched from [2, 10] with step 1, \{0.6, 0.3, 0.1, 0.01,0.001\} and \{0, 1e-3, 1e-4, 1e-5\}. Dropout, $T$ and $\lambda$ are searching from [0, 0.7], [1, 2] and [0, 1] with step 0.1, respectively. Notice that, for GLNN, we follow their paper and set the hidden embedding size as 4-times and 8-times wider than the base model on dataset Ogbn-arxiv and Ogbn-products. To eliminate randomness, we repeat each method three times and report the mean performance. The code is written in Python 3.9 and the operating system is Ubuntu 16.0. We use Pytorch 1.11.0 on CUDA 11.7 to train models on GPU. The inference time is evaluated on the CPU with batch size 500. All experiments are conducted on a machine with Intel(R) Xeon(R) CPUs (Gold 5120 @ 2.20GHz) and NVIDIA TITAN RTX GPUs with 24GB GPU memory.

\begin{table*}[pt]
\caption{{Inference comparison under base model SGC. Acceleration ratios between NAI and vanilla SGC are shown in brackets.}}
 \label{tab_SGC}
 \resizebox{0.95\textwidth}{!}{
\begin{tabular}{l|rrrrr|rrrrr}
\hline
             & \multicolumn{5}{c|}{PubMed}                                  & \multicolumn{5}{c}{Flickr}                                     \\ \hline
             & ACC (\%) & \# mMACs  & \#FP mMACs & Time (ms) & FP Time (ms) & ACC (\%) & \# mMACs   & \#FP mMACs & Time (ms)  & FP Time (ms) \\ \hline
SGC          & 80.00      & 244.3     & 243.5      & 393.0     & 340.6        & 49.43    & 2475.3     & 2471.2     & 2530.6     & 2381.8       \\
GLNN         & 79.43    & 0.7       & 0.0        & 7.6       & 0.0          & 44.39    & 4.2        & 0.0        & 11.0       & 0.0          \\
TinyGNN      & 79.61    & 658.9     & 658.2      & 420.3     & 413.7        & 46.80     & 8850.3     & 8846.1     & 1413.8     & 1412.1       \\
Quantization & 79.90     & 244.3     & 243.5      & 364.0     & 337.2        & 48.34    & 2475.3     & 2471.2     & 2482.2     & 2344.7       \\
NAI          & 79.97    & 7.1 (34)  & 1.3 (187)  & 18.4 (21) & 11.5 (30)    & 49.36    & 174.9 (14) & 148.3 (17) & 238.5 (11) & 143.4 (17)   \\ \hline
             & \multicolumn{5}{c|}{Ogbn-arxiv}                              & \multicolumn{5}{c}{Ogbn-products}                              \\ \hline
             & ACC (\%) & \# mMACs  & \#FP mMACs & Time (ms) & FP Time (ms) & ACC (\%) & \# mMACs   & \#FP mMACs & Time (ms)  & FP Time (ms) \\ \hline
SGC          & 69.36    & 895.1     & 887.8      & 1276.7    & 1034.2       & 74.24    & 32946.4    & 32939.7    & 68806.7    & 50628.6      \\
GLNN       & 54.83    & 108.0     & 0.0        & 19.4      & 0.0          & 63.12    & 337.0      & 0.0        & 238.9      & 0.0          \\
TinyGNN      & 67.31    & 294.6     & 287.2      & 523.7     & 522.1        & 71.33    & 3418.0     & 3411.3     & 1954.6     & 1948.2       \\
Quantization & 68.88    & 895.1     & 887.8      & 1223.4    & 1003.6       & 73.01    & 32946.4    & 32939.7    & 68726.0    & 50587.6      \\
NAI          & 69.25    & 83.5 (11) & 65.1 (14)  & 182.4 (7) & 60.6 (17)    & 73.70     & 583.2 (56) & 451.6 (73) & 923.2 (75) & 591.6 (86)   \\ \hline
\end{tabular}}
\end{table*}

\begin{figure*}[ht]
 \setlength{\abovecaptionskip}{2.mm}
  \centering
  \hspace{0mm}
  \subfloat[PubMed]{\includegraphics[width=0.24\textwidth]{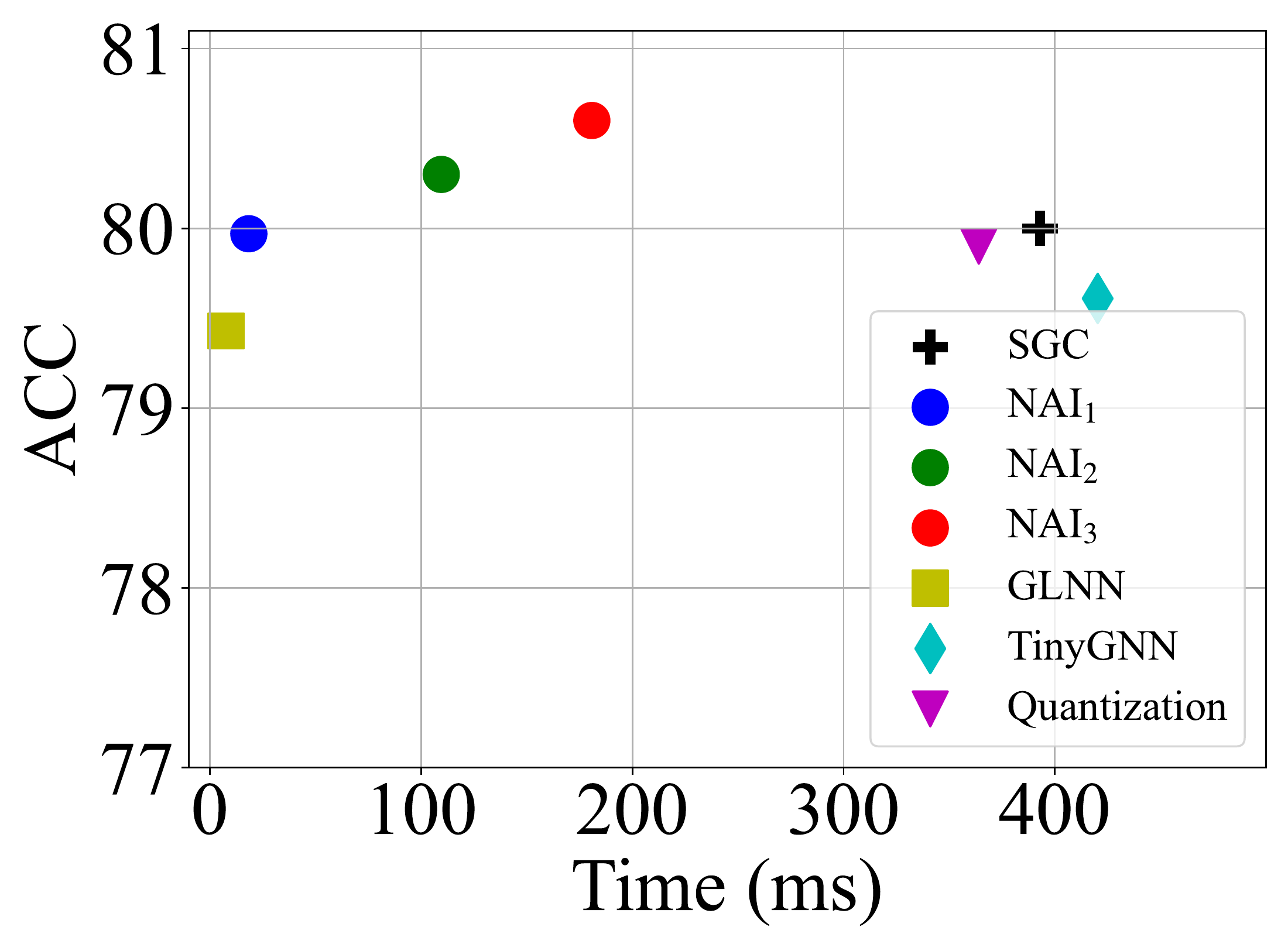}}
   \hspace{0mm}
    \subfloat[Flickr]{\includegraphics[width=0.24\textwidth]{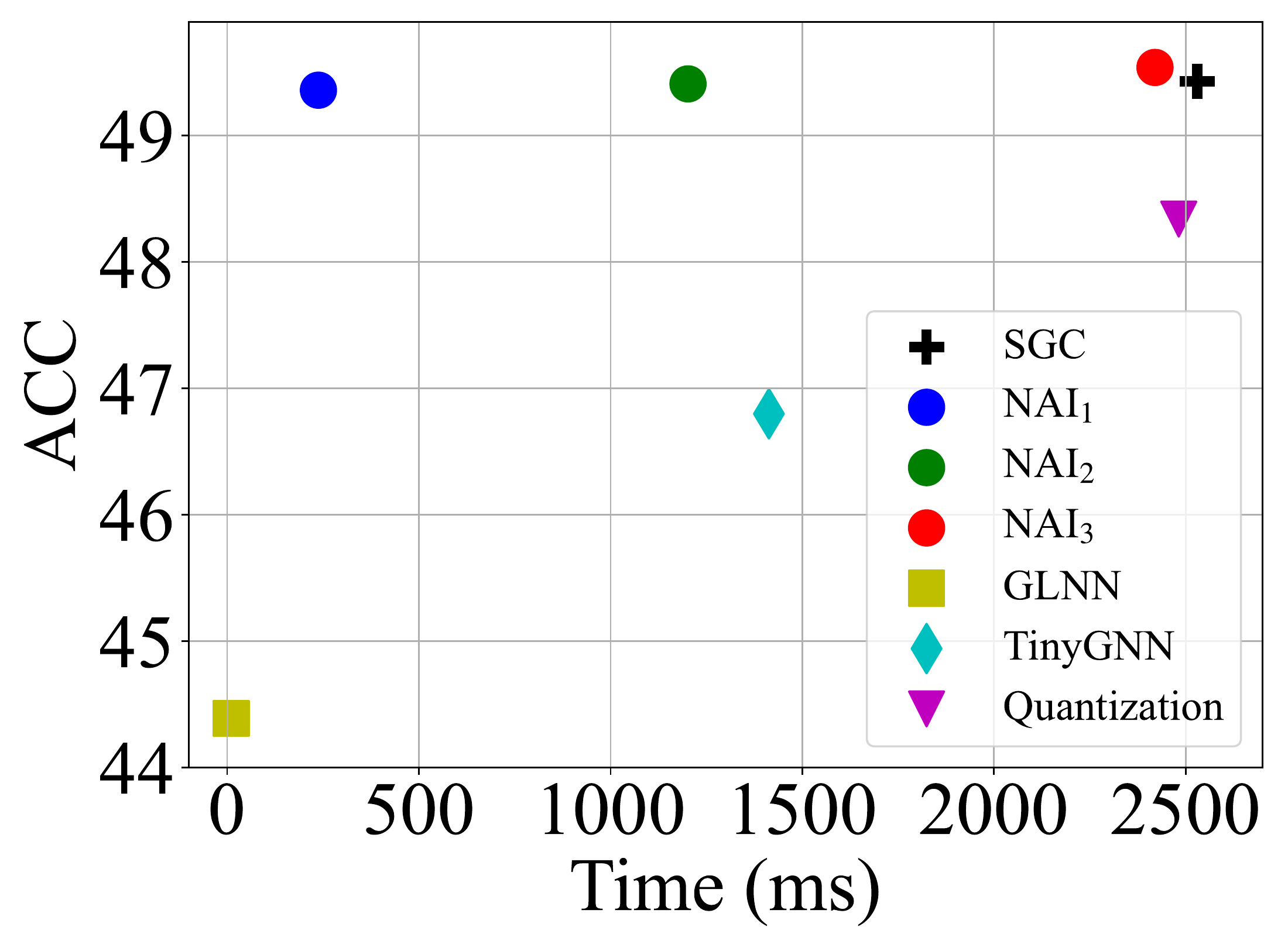}}
   \hspace{0mm}
    \subfloat[Ogbn-arxiv]{\includegraphics[width=0.24\textwidth]{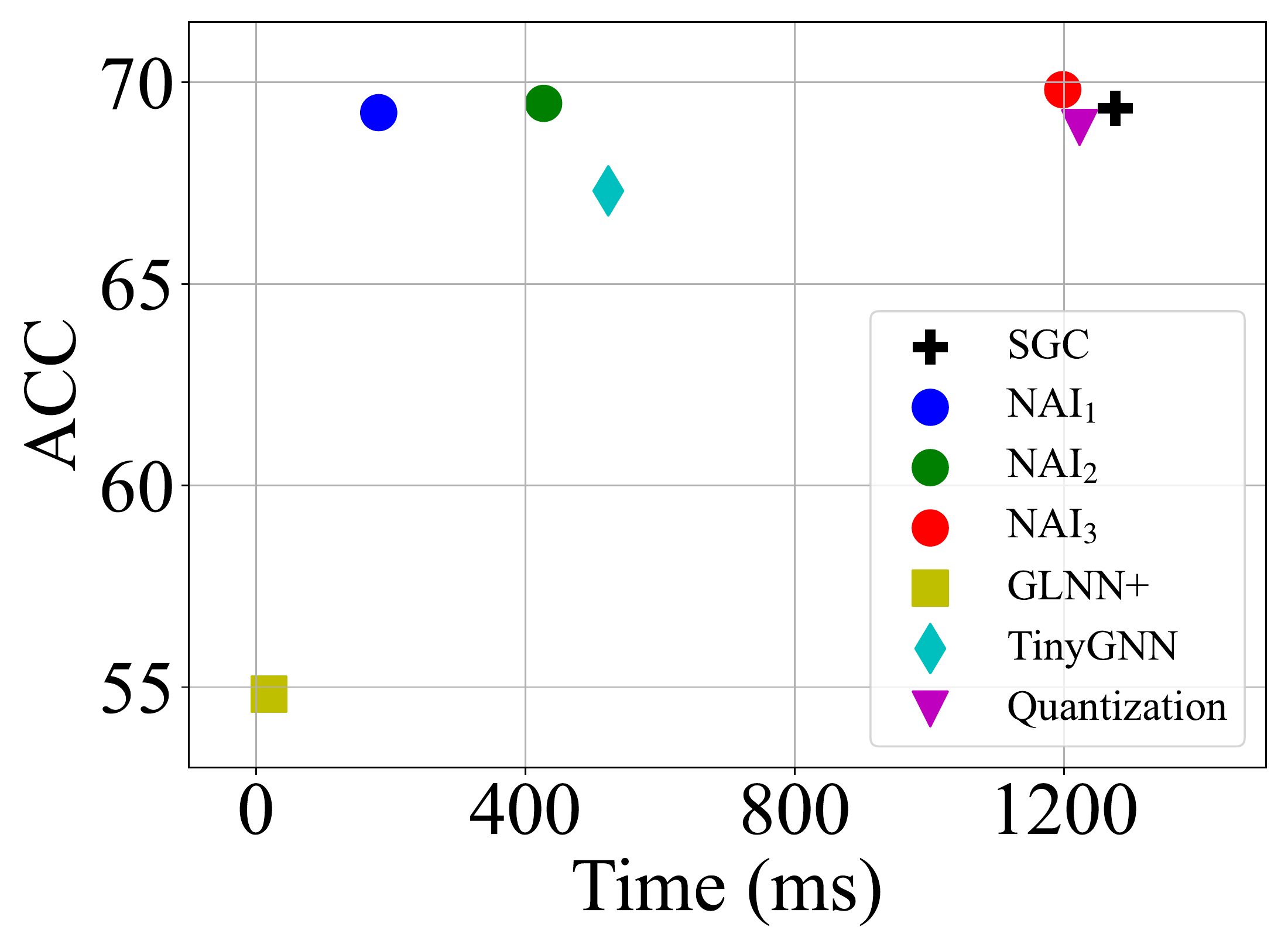}}
   \hspace{0mm}    
   \subfloat[Ogbn-products]{ \includegraphics[width=0.24\textwidth]{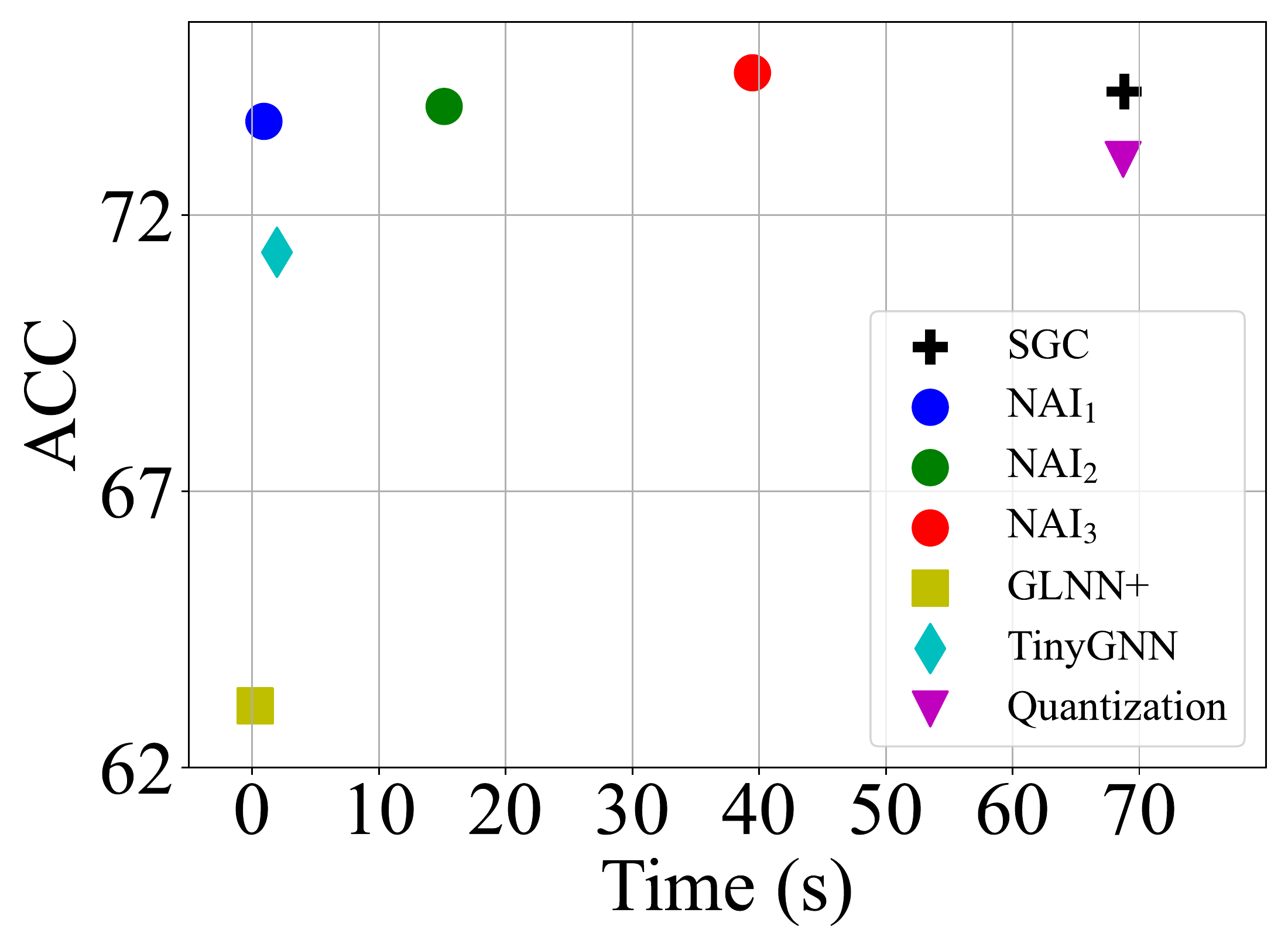}}  
  \caption{Accuracy and inference time comparison. NAIs with the subscript indicate 3 different settings.}
  \label{fig_sgc}
\end{figure*}

\begin{table*}[ht]
  \caption{Node distributions of NAI under different settings. The propagation order increases from 1 (left) to $k$ (right).}
  \label{fig_node_distribution}
\resizebox{0.65\textwidth}{!}{
\begin{tabular}{l|lll}
\toprule
              & \multicolumn{1}{c}{$\rm{NAI_1}$} & \multicolumn{1}{c}{$\rm{NAI_2}$} & \multicolumn{1}{c}{$\rm{NAI_3}$}        \\ \hline
PubMed        & {[}1000, 0, 0, 0, 0, 0, 0{]}  & {[}207, 319, 474, 0, 0, 0, 0{]} & {[}625, 224, 50, 25, 10, 21, 45{]}    \\
Fickr         & {[}76, 22237, 0, 0, 0, 0, 0{]}   & {[}0, 0, 1938, 20375, 0, 0, 0{]}  & {[}0, 392, 5580, 848, 85, 308, 15100{]} \\
Ogbn-arxiv    & {[}1849, 46754, 0, 0, 0{]}       & {[}0, 20528, 28075, 0, 0{]}      & {[}0, 16503, 12221, 1077, 18802{]}      \\
Ogbn-products & {[}0, 2213091, 0, 0, 0{]}        & {[}1086, 0, 2212005, 0, 0{]}     & {[}0, 1384, 239, 2211468, 0{]}            \\ \bottomrule
\end{tabular}}
\end{table*}

\subsection{Performance Comparison}
In Table \ref{tab_SGC}, we compare NAI with other baselines under the base model: SGC. For the NAI, we select the hyper-parameters that prioritize the inference speed.

From Table \ref{tab_SGC}, we observe that NAI has a great balance between accuracy and inference speed. As for ACC, NAI outperforms the Quantization method and achieves the least ACC loss compared to vanilla SGC. The maximum ACC drop among four datasets is controlled as 0.54\% on the Ogbn-products dataset. Although Quantization also shows great accuracy, it only saves the classification computation and could not help to reduce the computation from feature processing. For this reason, the maximum Time acceleration of Quantization is 1.08$\times$ on PubMed. Benefiting from removing the feature propagation in the inference procedure, GLNN has the smallest MACs and the fastest inference speed. However, for the same reason, GLNN could not generalize well for inductive settings as analyzed in their paper. Even with the increased embedding size, the accuracies on Ogbn-arxiv and Ogbn-products decrease significantly. This indicates that ignoring topological information severely impairs the prediction of unseen nodes. Moreover, NAI outperforms TinyGNN on all datasets. Although TinyGNN saves a part of the computation of feature propagation, the self-attention mechanism and linear transformation used in its peer-aware module cause a large 
number of extra computations. Especially in the dataset with high feature dimension, e.g., PubMed, the MACs and Time are much more than vanilla SGC. The peer-aware module takes up 98\% (405.2/413.7 ms) of the FP time in TinyGNN and directly results in higher latency. Compared with baselines, NAI accelerates inference significantly by controlling the FP MACs and achieves the 75$\times$ Time speedup and 86$\times$ FP Time speedup on Ogbn-products.

\begin{table*}[t]
\caption{{The ablation study on NAP under different $T_{max}$. The propagation order of node distribution increases from 1 to 5.}}
 \label{tab_exitnode}
 \resizebox{0.8\linewidth}{!}{
\begin{tabular}{r|l|rrl|rrl}
\toprule
                   & \multicolumn{1}{c|}{} & \multicolumn{3}{c|}{Ogbn-arxiv}                                     & \multicolumn{3}{c}{Ogbn-products}                                \\ \hline
$T_{max}$          & Method                & \multicolumn{1}{r}{ACC (\%)} & Time (ms) & Node distribution                       & \multicolumn{1}{r}{ACC (\%)} & Time (ms) & Node distribution                   \\ \hline
\multirow{2}{*}{2} & NAI w/o NAP           & 69.16                   & 202.7  & {[}0, 48603, 0, 0, 0{]}            & 73.70                   & 923.2   & {[}0, 2213091, 0, 0, 0{]}      \\
                   & NAI                   & 69.25                   & 182.4  & {[}1849, 46754, 0, 0, 0{]}         & 73.70                   & 923.2   & {[}0, 2213091, 0, 0, 0{]}     \\ \hline
\multirow{2}{*}{3} & NAI w/o NAP           & 69.38                   & 454.2  & {[}0, 0, 48603, 0, 0{]}            & 73.95                   & 17121.5 & {[}0, 0, 2213091, 0, 0{]}      \\
                   & NAI                   & 69.48                   & 427.4  & {[}0, 20528, 28075, 0, 0{]}        & 73.97                   & 15146.1 & {[}1086, 0, 2212005, 0, 0{]}   \\ \hline
\multirow{2}{*}{4} & NAI w/o NAP           & 69.26                   & 889.3  & {[}0, 0, 0, 48603, 0{]}            & 74.57                   & 42232.2 & {[}0, 0, 0, 2213091, 0{]}      \\
                   & NAI                   & 69.52                   & 816.6  & {[}0, 30303, 5898, 12402, 0{]}     & 74.58                   & 39474.8 & {[}0, 1384, 239, 2211468, 0{]} \\ \hline
\multirow{2}{*}{5} & NAI w/o NAP           & 69.36                   & 1296.4 & {[}0, 0, 0, 0, 48603{]}            & 74.24                   & 68938.8  & {[}0, 0, 0, 0, 2213091{]}     \\
                   & NAI                   & 69.82                   & 1198.9 & {[}0, 16503, 12221, 1077, 18802{]} & 74.58                   & 67523.2 & {[}0, 0, 0, 2213068, 23{]}     \\ \bottomrule
\end{tabular}}
\end{table*}

Besides the speed-first results in Table \ref{tab_SGC}, NAI allows users to choose more accurate results based on the latency constraints. Figure \ref{fig_sgc} shows the trade-off between accuracy and inference time in different hyper-parameter settings. We select 3 typical settings for each dataset, which are denoted as "$\rm{NAI_1}$", "$\rm{NAI_2}$" and "$\rm{NAI_3}$", respectively. Note that "$\rm{NAI_1}$" is the speed-first setting in Table \ref{tab_SGC}. From Figure \ref{fig_sgc}, NAIs achieve the highest classification accuracy and even superior to vanilla SGC. This is due to that NAP mitigates the over-smoothing problem and Inception Distillation enhances the classifiers (Table \ref{tab_exitnode} and \ref{tab_ablation} in next subsection evaluate their impacts). 
For example, on Flickr, $\rm{NAI_3}$ achieves more accurate results while spending a similar inference time with SGC, and $\rm{NAI_2}$ further accelerates $\rm{NAI_3}$ by 2.1$\times$ with little accuracy drop. 
Moreover, Table \ref{fig_node_distribution} shows the detailed test node distribution over different datasets and hyper-parameter settings, i.e., the number of nodes with the different propagation orders. The order increases from 1 (left) to $k$ (right).
From Table \ref{fig_node_distribution}, we observe that most of the nodes of $\rm{NAI_2}$ on Flickr adopt 4-th order propagated features. This successfully reduces the number of supporting nodes and saves the computation of the feature propagation. To get the best accuracy, $\rm{NAI_3}$ makes full use of each classifier, and the propagation orders of tested nodes are various. As for the $\rm{NAI_1}$ on Ogbn-products, all nodes adopt the 2nd order propagated features to trade off the inference speed and accuracy. It demonstrates the flexibility of NAI, and the fixed propagation order used in classic GNNs is the special case of our proposed method.

\subsection{Ablation Study}

To thoroughly evaluate our method, we provide ablation studies on: (1) Node-Adaptive Propagation; (2) Inception Distillation.

Table \ref{tab_exitnode} shows the performance of NAI and NAI without NAP under different hyper-parameter settings on Ogbn-arxiv and Ogbn-products. Their maximum propagation orders are $k=5$, and $T_{max}=1$ is omitted due to the same inference results. Under the same  $T_{max}$, the selection of hyper-parameters of NAI prioritizes the accuracy. Note that the accuracies of "NAI w/o NAP" do not grow monotonically with $T_{max}$ because the Inception Distillation enhances the classifiers independently. Comparing NAI with "NAI w/o NAP" under the same $T_{max}$, accuracies are all improved with less inference latency. To achieve a fast inference speed under the same $T_{max}$, tested nodes adopt various propagation orders, e.g., there are 62.3\% nodes propagated twice when $T_{max}$=4 on Ogbn-arxiv, contributing to both accuracy improvement and computation saving. These experimental results illustrate that NAP provides more flexible inference patterns and could mitigate the over-smoothing problem successfully.

\begin{table}[h]
\caption{{The ablation study on the Inception Distillation. Accuracies (\%) are averaged over 3 runs.}}
 \label{tab_ablation}
  \resizebox{0.9\linewidth}{!}{
\begin{tabular}{l|rrrrrr}
\toprule
                      & PubMed     & Flickr     & Ogbn-arxiv & Ogbn-products \\ \hline
NAI w/o ID                 & 75.96 & 40.86 & 65.54 & 70.17    \\
NAI w/o ON & 79.59 & 44.41 & 65.91 & 70.28    \\
NAI w/o OFF & 79.58 & 42.81 & 66.08 & 70.37    \\
NAI  & 79.97 & 44.85 & 66.10 & 70.49    \\ \bottomrule
\end{tabular}}
\end{table}

Besides NAP, Inception Distillation is designed to explore multi-scale knowledge and improve the inference accuracy. We evaluate the accuracy of $f^{(1)}$, which has the worst performance among classifiers, to show the effectiveness of each component in Inception Distillation. Table \ref{tab_ablation} displays the results of NAI without Inception Distillation ("w/o ID"), NAI without offline distillation ("w/o OFF"), NAI without online distillation ("w/o ON") and NAI. First, the online distillation explores multi-scale reception features and constructs a more powerful teacher via self-attention mechanism, contributing to improvements on all datasets when comparing NAI with NAI w/o ON. For example, when ignoring the online distillation, the accuracy of NAI will drop 0.44\% on Flickr. Besides, offline distillation provides a solid foundation for online distillation. With more accurate classifiers, the ensemble teacher will be more expressive and powerful, which could provide higher-quality supervision signals. The classification results will decrease on all datasets when offline distillation is removed. With the help of offline distillation, the accuracy of online distillation has a 2.04\% increase on the dataset Flickr. These results indicate that both offline and online distillation are essential to NAI.

\subsection{Generalization}
In addition to SGC, our proposed method can be applied to any linear-propagation based GNNs. We test the generalization ability of NAI by deploying NAI on $\mathrm{S^2GC}$, SIGN and GAMLP on Flickr. The hyper-parameters, including the classifier structure, are searched to get the best performance for each base model. The propagation orders for $\mathrm{S^2GC}$, SIGN and GAMLP are 10, 5 and 5, respectively. 

\begin{table*}[ht]
  \setlength{\abovecaptionskip}{3.mm}
\caption{{Inference comparison under different base models on Flickr. ACC is evaluated in percentage. Time and FP Time are evaluated in millisecond. Acceleration ratios between NAI and vanilla GNNs are shown in brackets.}}
 \label{tab_basemodel}

  \resizebox{\textwidth}{!}{
\begin{tabular}{l|rrrrr|rrrrr|rrrrr}
\toprule
             & \multicolumn{5}{c|}{$\mathrm{S^2GC}$}                                 & \multicolumn{5}{c|}{SIGN}                                 & \multicolumn{5}{c}{GAMLP}                                \\ \hline
             & ACC      & \#mMACs & \#FP mMACs & Time    & FP Time      & ACC     & \#mMACs & \#FP mMACs & Time       & FP Time   & ACC    & \#mMACs & \#FP mMACs & Time    & FP Time  \\ \hline
Vanilla GNN    & 50.08 & 3897.8 & 3889.2    & 3959.5 & 3717.6 &  51.00 & 1574.9 & 1526.8    & 1667.1  & 1569.0  & 51.18 & 1594.8 & 1590.6    & 1759.6  & 1657.6 \\
GLNN       & 46.59 & 8.6    & 0       & 9.5     & 0    &  46.84 & 8.1   & 0       & 7.8      & 0     & 46.99 & 8.2   & 0      & 7.2    & 0  \\
TinyGNN     & 46.89 & 8855.1 & 8846.5    & 1366.7 & 1355.0   &   47.21 & 8862.2 & 8846.1    & 1356.1 & 1345.9   & 47.40 & 8875.8 & 8873.7    & 1389.1 & 1381.8    \\
Quantization & 49.10 & 3897.8 & 3889.2    & 3946.9 & 3714.6    & 45.87 & 1574.9 & 1526.8    & 1654.3  & 1565.0  & 50.81 & 1594.8 & 1590.6    & 1701.6  & 1650.8  \\
NAI    & 48.94 & 120.1 (32) & 89.0 (44)   & 149.9 (26)   & 86.3 (43)   &  51.02 & 135.0 (12)  & 112.5 (14)     & 170.4 (10)   & 78.7 (20)   & 50.89 & 150.0 (11)  & 124.9 (13)     & 220.1 (8)   & 133.3 (12)    \\ \bottomrule
\end{tabular}}
\end{table*}

 The accuracy and inference time results are shown in Table \ref{tab_basemodel}. NAI consistently outperforms the other baselines when considering both accuracy and inference speedup. Compared to GLNN, NAI can improve the accuracy for 2.35\%, 4.18\% and 3.90\% on $\mathrm{S^2GC}$, SIGN and GAMLP, respectively. Although the attention mechanism used in TinyGNN requires a large number of MACs, the feature propagation is more time-consuming on Flickr and the acceleration ratios for different base models are ranging from 1.2$\times$ to 2.9$\times$ compared with vanilla GNNs. Quantization achieves the smallest accuracy loss but the acceleration ratio is limited. When applying NAI to $\mathrm{S^2GC}$, SIGN and GAMLP, the FP Time can be accelerated by $43\times$, $20\times$ and $12\times$. Considering the other computations, i.e., the computation of stationary state and classification, the corresponding inference time are accelerated by $26\times$, $10\times$ and $8\times$.

\subsection{Parameter Sensitivity Analysis}
\begin{figure}[ht]
  \centering
  
\setlength{\abovecaptionskip}{0.cm}
  \hspace{-3mm}
  \includegraphics[width=0.327\linewidth]{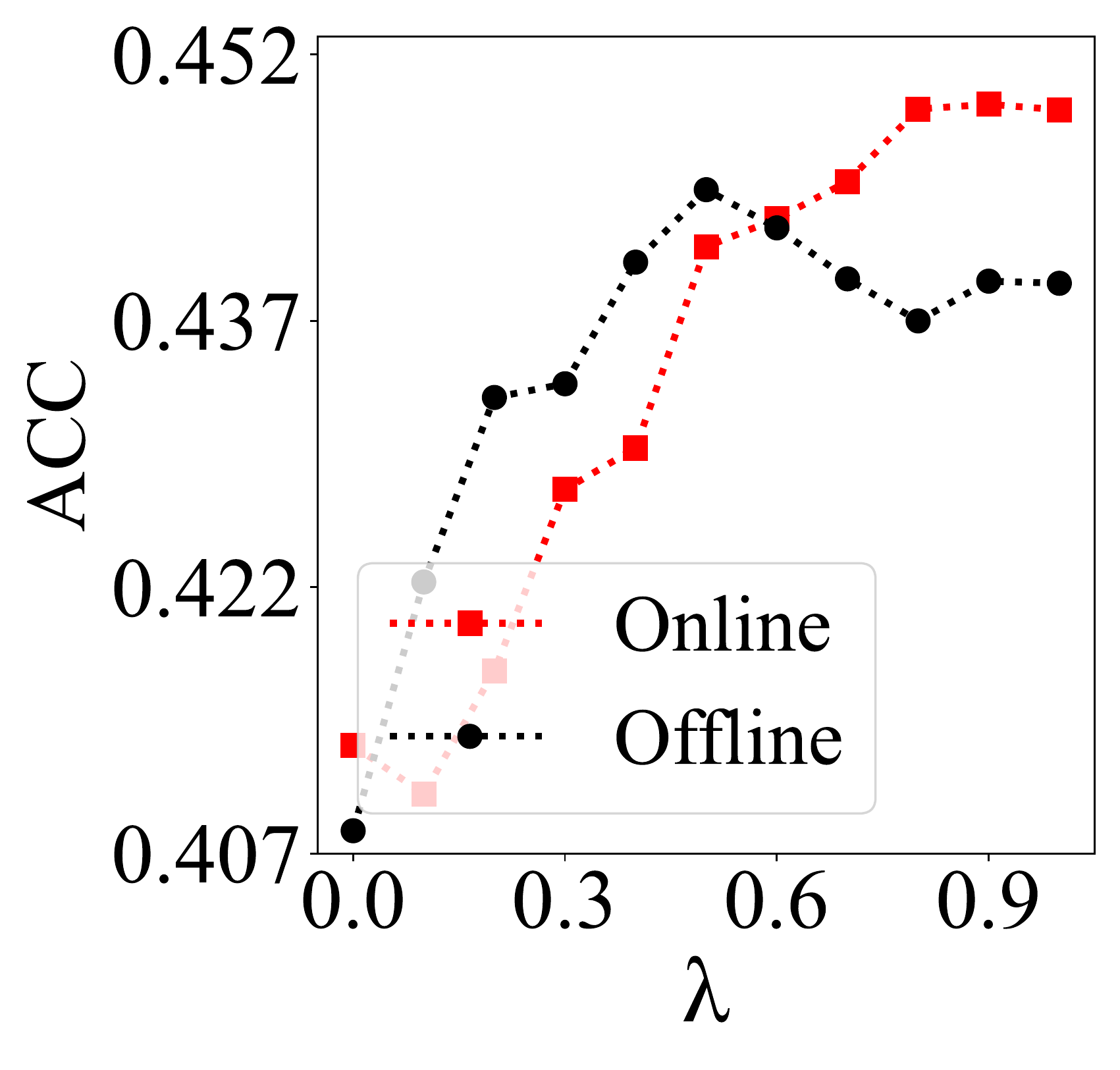}
   \hspace{-2mm}
    \includegraphics[width=0.327\linewidth]{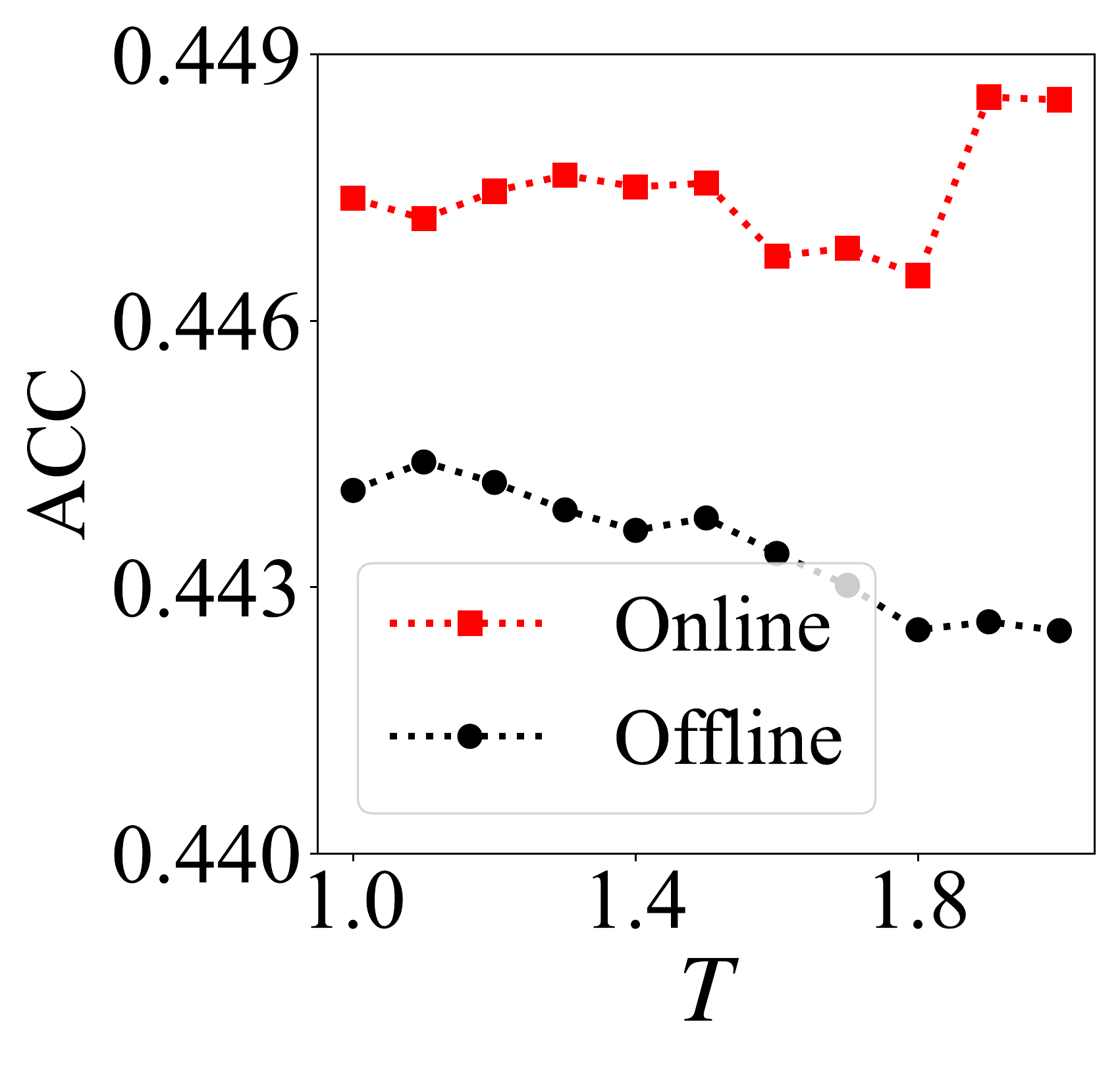}
   \hspace{-2mm}
    \includegraphics[width=0.327\linewidth]{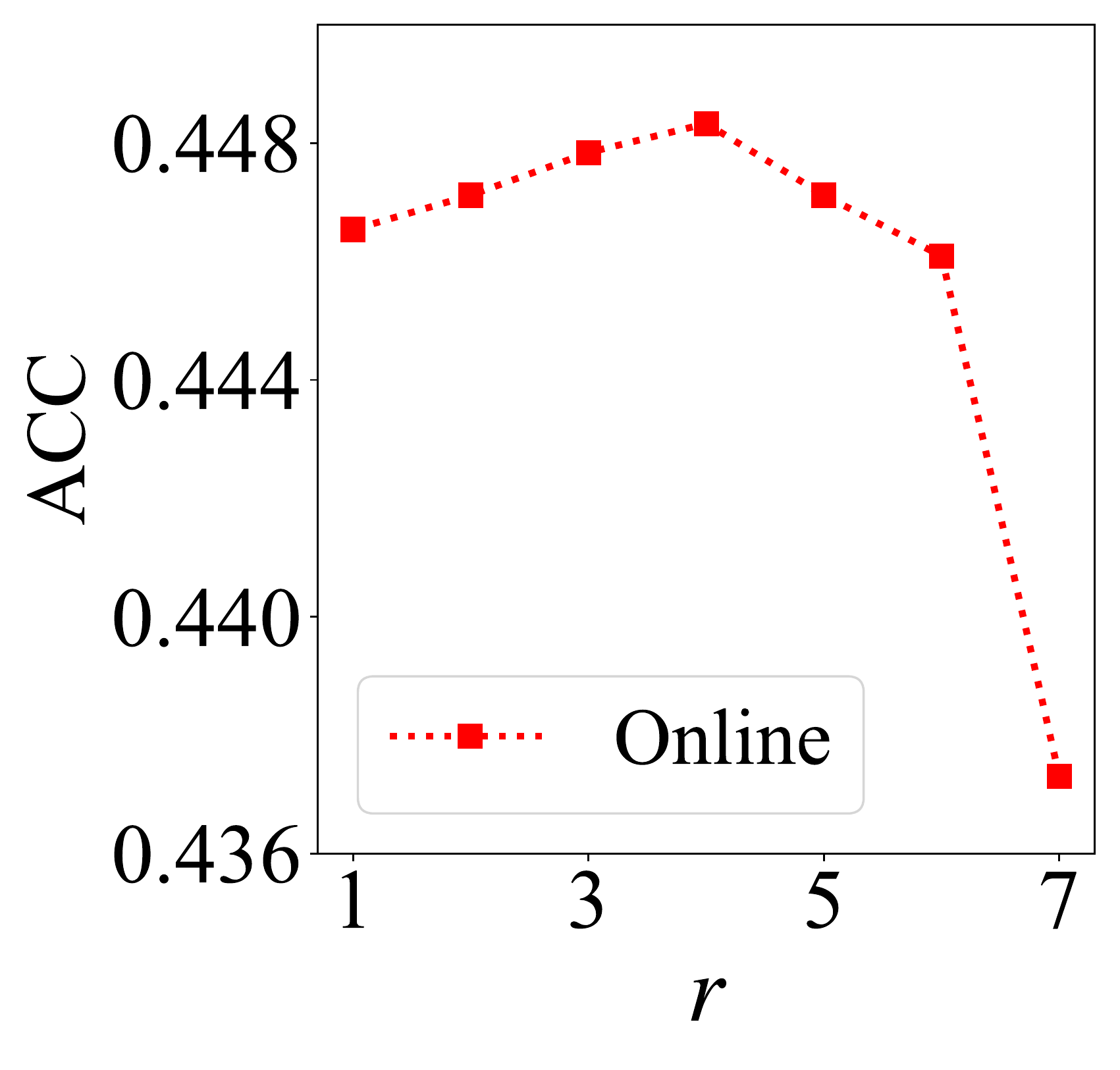}
  \caption{Parameter sensitivity results on Flickr.}
  \label{fig_para}
\end{figure}

Temperature $T$ and weight $\lambda$ are two influential hyper-parameters for Inception Distillation. Moreover, the ensemble number $r$ controls the teacher quality in online distillation. To analyze the influence of these hyper-parameters, 
we conduct the experiment on Flickr and the base model is SGC. The classification performances of $f^{(1)}$ in terms of hyper-parameters are shown in Figure \ref{fig_para}.

Firstly, $\lambda$ is quite important which could significantly affect the classification result. For example, $\lambda$ for online distillation should be controlled between 0.8 and 1 to get better performance. This indicates that the supervision provided by the ensemble teacher is more important than the hard label. In contrast, $\lambda$ for offline distillation should be selected carefully to balance two losses. Following the increase of $T$, the performance of online distillation decreases first and then increases. Thus, limiting $T$ to a larger value and using softer labels works best. The offline distillation results in terms of $T$ show that decreasing temperature could help enhance the classification performance. $T$ should be controlled in the range of [1, 1.2]. Finally, the results in terms of $r$ show that increasing the number of combined models could help enhance the classification performance. But it also introduces more unreliable labels in model training. Especially when introducing the low quality labels from $f^{(1)}$, the classification result drops rapidly. To sum up, Inception Distillation gets stable and high classification performances when $\lambda$ ranges from 0.5 to 1. Softer labels and an appropriate ensemble number should be applied to online distillation for better performance.

\section{Related Works}
To deploy the model on large-scale graphs, researchers propose various techniques to accelerate training and inference, which can be categorized into model perspective and algorithm perspective. 

From the model perspective, scalable GNNs mainly contain sampling-based models and linear propagation-based models. Besides the models studied in this paper, sampling-base models can be divided into three categories according to sampling methods: node-wise \citep{hamilton2017inductive, chen2017stochastic}/ layer-wise \citep{DBLP:conf/iclr/ChenMX18, huang2018adaptive, zou2019layer}/ graph-wise \citep{chiang2019cluster, DBLP:conf/iclr/ZengZSKP20} sampling. 
Although sampling-based GNNs mitigate the neighbor explosion problem by restricting the number of neighbors, they are greatly influenced by sampling quality and suffer from the high variance problem when applied to inference. 

From the algorithm perspective, acceleration methods include pruning, quantization and knowledge distillation (KD). Pruning methods designed for GNNs \citep{zhou2021accelerating} reduce the dimension of embeddings in each hidden layer to save the computation. Quantization \citep{tailor2020degree} uses low-precision integer arithmetic during inference to speed up the computation. However, these two kinds of methods concentrate on reducing the computation of feature transformation and classification, and raw features are preserved to avoid performance degradation. This limits the acceleration performance considering that feature propagation accounts for the most proportion of runtime. KD aims to train a light-weight model which has a similar performance to the teacher model. Most KD methods for GNNs try to enhance the student performance by introducing high-order structural information because the receptive field is bound to the number of GNNs layers \citep{yang2020distilling, jing2021amalgamating, yang2021extract}. 
Besides, GraphAKD \citep{he2022compressing} leverages adversarial training to decrease the discrepancy between teacher and student. ROD \citep{zhang2021rod} uses multiple reception field information to provide richer supervision signals for sparsely labeled graphs. RDD \citep{zhang2020reliable} defines the node reliability and edge reliability to make better use of high-quality data.
Different from the above works which concentrate on improving the performance of a single model, the Inception Distillation in NAI focuses on multi-scale knowledge transfer and boosts the performance for multiple students.

Another type of related work for inference acceleration is the early exiting technique, which allows samples to exit early from the backbone network to accelerate the inference and has been widely used in CV \citep{teerapittayanon2016branchynet, phuong2019distillation} and NLP \citep{xin2020deebert, liu2020fastbert, zhou2020bert}.
As the first work (to our best knowledge) utilizing the similar idea in graph representation learning, NAI mainly focuses on reducing the computational redundancies of feature propagation. Compared with works in CV and NLP, the exiting criteria of NAI takes the topological information of graph into account instead of referring to the prediction results. This allows each node to be predicted once only and saves the computation cost by repeated prediction in other works. In addition, benefiting from non-parameterized feature propagation, the training procedure of NAI is more stable and flexible. In other works, the performance of later exits relies on the embeddings of former exits. This makes that all exits and backbone have to be trained together and the loss functions need to be carefully designed.

\section{Conclusion}
We present Node-Adaptive Inference (NAI), a general inference acceleration method for scalable GNNs. NAI can successfully reduce the redundancy computation in feature propagation and achieve adaptive node inference with personalized propagation orders. With the help of Inception Distillation, NAI exploits multi-scale reception field knowledge and compensates for the potential inference accuracy loss. Extensive experiments on large-scale graph datasets verified that NAI has high acceleration performance, good generalization ability and the flexibility for different latency constraints. NAI drives the industrial applications of scalable GNNs, especially in streaming and real-time inference scenarios.


\bibliographystyle{ACM-Reference-Format}
\bibliography{ref}


\end{document}